\pdfoutput=1

\documentclass[11pt,a4paper]{article}
\usepackage[hyperref]{acl2021}
\usepackage{times}
\usepackage{latexsym}

\usepackage{microtype}

\usepackage{times}
\usepackage{latexsym}

\usepackage[T1]{fontenc}

\usepackage[utf8]{inputenc}

\usepackage{graphicx}
\usepackage{url}
\usepackage[english]{babel}
\usepackage{booktabs}
\usepackage{multirow}
\usepackage{footnote}
\usepackage{amssymb}
\usepackage{enumitem}
\usepackage{color, colortbl}
\makesavenoteenv{tabular}
\makesavenoteenv{table}
\usepackage{footmisc}
\usepackage[most]{tcolorbox}
\usepackage{makecell}
\usepackage{tablefootnote}

\usepackage{newfloat,caption,float}
\DeclareFloatingEnvironment[
  fileext = loe,
  listname = Examples,
  name = Example,
  placement = H,
  within = none
  ]{example}
\usepackage{etoolbox}
\AfterEndEnvironment{example}{\vskip-2ex}

\makeatletter
\renewcommand*{\@fnsymbol}[1]{\ensuremath{\ifcase#1\or \dagger\or \ddagger\or
    \mathsection\or \mathparagraph\or \|\or **\or \dagger\dagger
    \or \ddagger\ddagger \else\@ctrerr\fi}}
\makeatother

\makeatletter
\newcommand{\printfnsymbol}[1]{%
  \textsuperscript{\@fnsymbol{#1}}%
}
\makeatother

\newcommand{\DEVELOPMENT}{0} 
\usepackage{ifthen}
\ifthenelse{\DEVELOPMENT = 1}{
	\newcommand{\tz}[1]{\textcolor{purple}{\textbf{TZ:} #1}}	
    \newcommand{\fz}[1]{\textcolor{teal}{\textbf{FZ:} #1}}
    \newcommand{\mh}[1]{\textcolor{orange}{\textbf{MH:} #1}}
}{
	\newcommand{\tz}[1]{}		
	\newcommand{\fz}[1]{}	
	\newcommand{\mh}[1]{}
}

\title{A Legal Approach to Hate Speech --\\ Operationalizing the EU's Legal Framework against the Expression of Hatred as an NLP Task}

\author{Frederike Zufall\textsuperscript{1}, Marius Hamacher\textsuperscript{2}, Katharina Kloppenborg\textsuperscript{2}, Torsten Zesch\textsuperscript{2}\\
     \textsuperscript{1} Max Planck Institute for Research on Collective Goods, Bonn, Germany;\\ Waseda Institute for Advanced Study, Waseda University, Tokyo, Japan\\
\textsuperscript{2} Language Technology Lab, 
	University of Duisburg-Essen, Germany \\
\tt{{\small zufall@coll.mpg.de, katharina.kloppenborg@cri-paris.org, }} \\
\tt{{\small  \{marius.hamacher, torsten.zesch\} @uni-due.de}} }

\date{}

\begin{document}
\maketitle

\begin{abstract}
We propose a 'legal approach' to hate speech detection by operationalization of the decision as to whether a post is subject to criminal law into an NLP task.
Comparing existing regulatory regimes for hate speech, we base our investigation on the European Union's framework as it provides a widely applicable legal minimum standard.
Accurately judging whether a post is punishable or not usually requires legal training.
We show that, by breaking the legal assessment down into a series of simpler sub-decisions, even laypersons can annotate consistently.
Based on a newly annotated dataset, our experiments show that directly learning an automated model of punishable content is challenging.
However, learning the two sub-tasks of `target group' and `targeting conduct' instead of an end-to-end approach to punishability yields better results.
Overall, our method also provides decisions that are more transparent than those of end-to-end models, which is a crucial point in legal decision-making.
\end{abstract}

\section{Introduction}
Social media provides the platform for the expression of opinions along with their widespread dissemination. Unrestricted freedom of expression, however, bears the risk of harming certain groups of people - rendering the regulation of hate speech a potential instrument against discrimination. To do so at scale, automated detection systems are required to aid the moderation process.
While research on hate speech detection is well-established, defining `hate speech' remains challenging.
Datasets encode all kinds of (partly incompatible) notions of hatefulness or offensiveness \citep{Schmidt2017, Fortuna2018, Poletto2020, Fortuna2021} that make it difficult to decide which postings would justify restricting freedom of speech through deletion.
Ultimately, a subset of especially hateful content can be considered punishable by law and thus would not fall under the legal right to freedom of expression. 
As there exist competing legal standards for the regulation of hateful expressions, the selection requires discussion.

\paragraph{Competing Legal Standards}
\label{sec:legal-standards}
On the international level, Article~4 of the `International Convention on the Elimination of All Forms of Racial Discrimination (ICERD)'\footnote{General Assembly resolution 2106 (XX) of 21 Dec 1965.} binds the signatory states to punish incitement to racial discrimination against any race or group of persons of another colour or ethnic origin by their respective national law.
However, the convention does not cover discrimination based on religion and is limited in its legal effect, as various states have made reservations.
This is especially the case for the U.S., where the expression of hatred toward any group is constitutionally widely protected by the Free Speech Clause of the First Amendment \cite{Fisch2002}. 
Consequently, as US law does not provide for any legal provision prohibiting hate speech as an act of speech, it cannot serve as a base for a detection system.

In Europe, however, the prevention of discrimination against and segregation of a target group (thereby ensuring the members' acceptance as equal in a society) is considered such an important prerequisite for democracy that it may justify the restriction of free speech. 
The Council of Europe has set up an additional protocol to the `Convention on Cybercrime', concerning the criminalization of acts of a racist and xenophobic nature committed through computer systems.\footnote{ETS No. 189, 28.01.2003.}
However, the Protocol has not been ratified or even signed by all Member States of the Council of Europe and is subject to several reservations.\footnote{Bulgaria, Hungary, Ireland, the Russian Federation and the U.K., for instance, did not sign the Protocol. Countries like Austria, Belgium, Italy, Sweden, Switzerland and Turkey signed, but did not (yet) ratify it.}

Legally and practically more relevant is the following instrument: the European Union (EU) has, after long debate, set up a common regime with a \textit{Framework Decision}\footnote{Council Framework Decision 2008/913/JHA of 28 November 2008 on combating certain forms and expressions of racism and xenophobia by means of criminal law. In the remainder of this paper, we shall refer to this as `EU law' or `EU Framework Decision' for simplification.} that fully binds all of its Member States to make incitement to hatred or violence a punishable criminal offense. 
The framework also affects U.S.\  social-media platforms as long as the offender or the material hosted is located within the EU. 
Its importance has also been emphasized by the `EU Code of conduct on countering illegal hate speech online' that the EU Commission agreed with IT companies like Facebook, Twitter, and Youtube.\footnote{\url{https://ec.europa.eu/newsroom/just/document.cfm?doc_id=42985}}
Furthermore, the EU's new proposal of a Digital Services Act aims to create new obligations for large online platforms regarding illegal content.\footnote{Proposal of 15.12.2020, COM(2020) 825 final.}
The regulation would not only be directly applicable in all EU Member States, but also apply to providers established outside the EU if they provide their services to recipients in the Union.
Hence, the EU Framework Decision not only provides a minimum standard for handling hate speech by criminal law, but it is also the regime that -- in connection with the new Digital Services Act -- would trigger the broadest regulatory obligations for large platform providers inside and outside the EU.


\begin{figure}[t]
    \centering
    \includegraphics[scale=0.35]{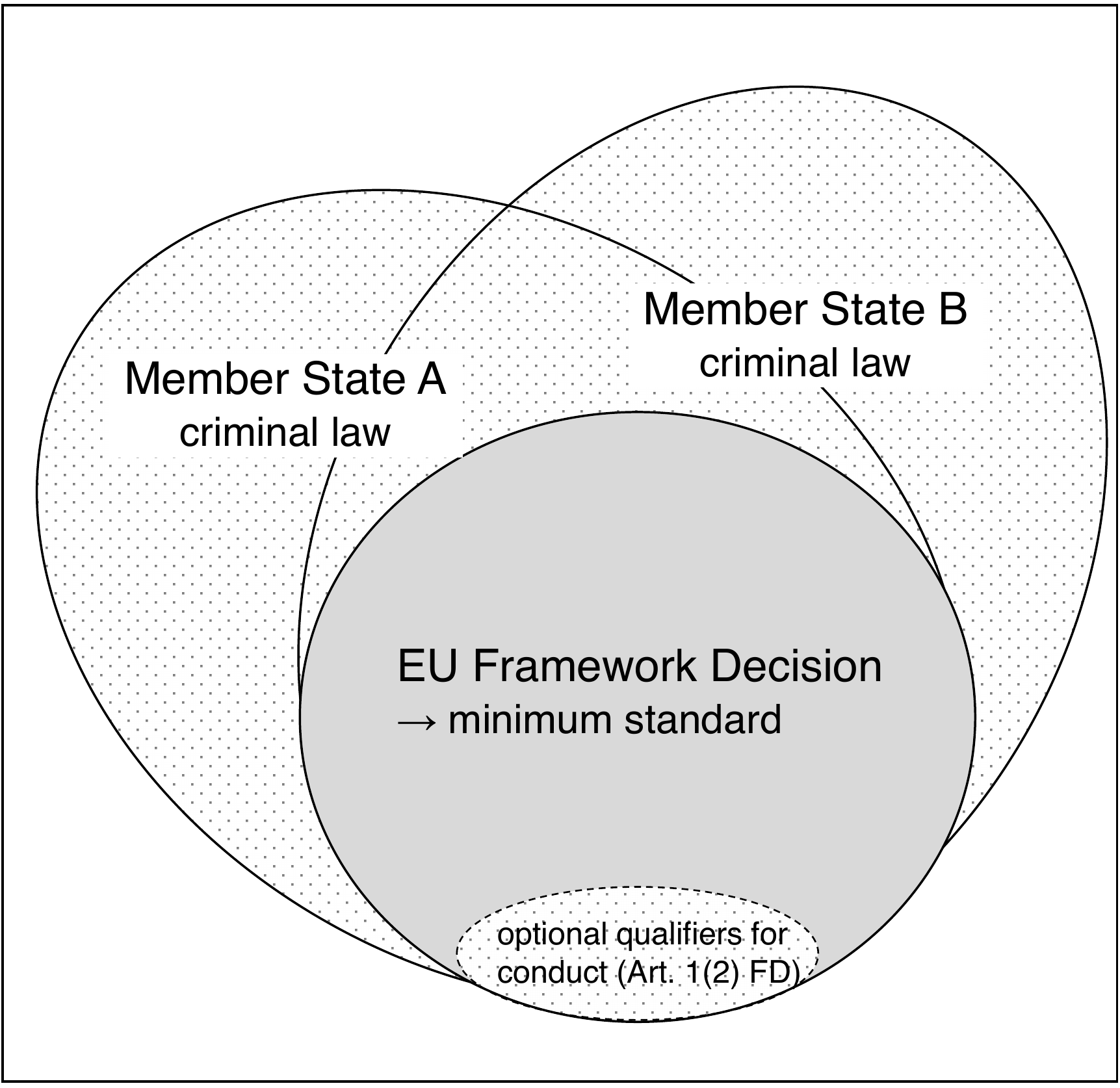}
\caption{Scope of the EU Framework's legal standard. It defines a common core of punishable offenses.}
\label{fig:EU_MS}
\end{figure} 

As Figure~\ref{fig:EU_MS} shows, each Member State may still go beyond the framework's minimum requirements and define higher standards. 
Germany, for instance, provides for a broader definition of the possible protected target group by including `sections of the population', e.g.\ refugees otherwise not being covered as they cannot clearly be distinguished by race, ethnic, or national origin.
However, the Framework Decision allows member states to make the incrimination depend on additional requirements.


Based on all these considerations, the Framework Decision's minimum standard may stand in for a general legal approach to hate speech and serve as the basis of our further studies.

\paragraph{Contributions}
In this paper, we translate the legal framework as defined in the EU Framework Decision 2008/913/JHA into a series of binary decisions.
We show that the resulting annotation scheme can be used by laypeople to reliably produce a legal evaluation of posts that is comparable to those of legal experts, making dataset generation for this task feasible.
Based on the resulting dataset, we experiment with directly learning an automated model of punishable content.
The discouraging results of the end-to-end approach and ethical considerations lead us to proposing two sub-tasks instead: `target group' and `targeting conduct' detection. We show that the sub-tasks can be more reliably learned and also provide for better explainability and higher transparency, which is a crucial point in legal decision-making.
We make our dataset and models publicly available to foster future research in that direction.

\section{Operationalizing Legal Assessment}
\label{sec:operationalizing}
\begin{figure}[t]
    \scriptsize
    \centering
 \includegraphics[scale=0.4]{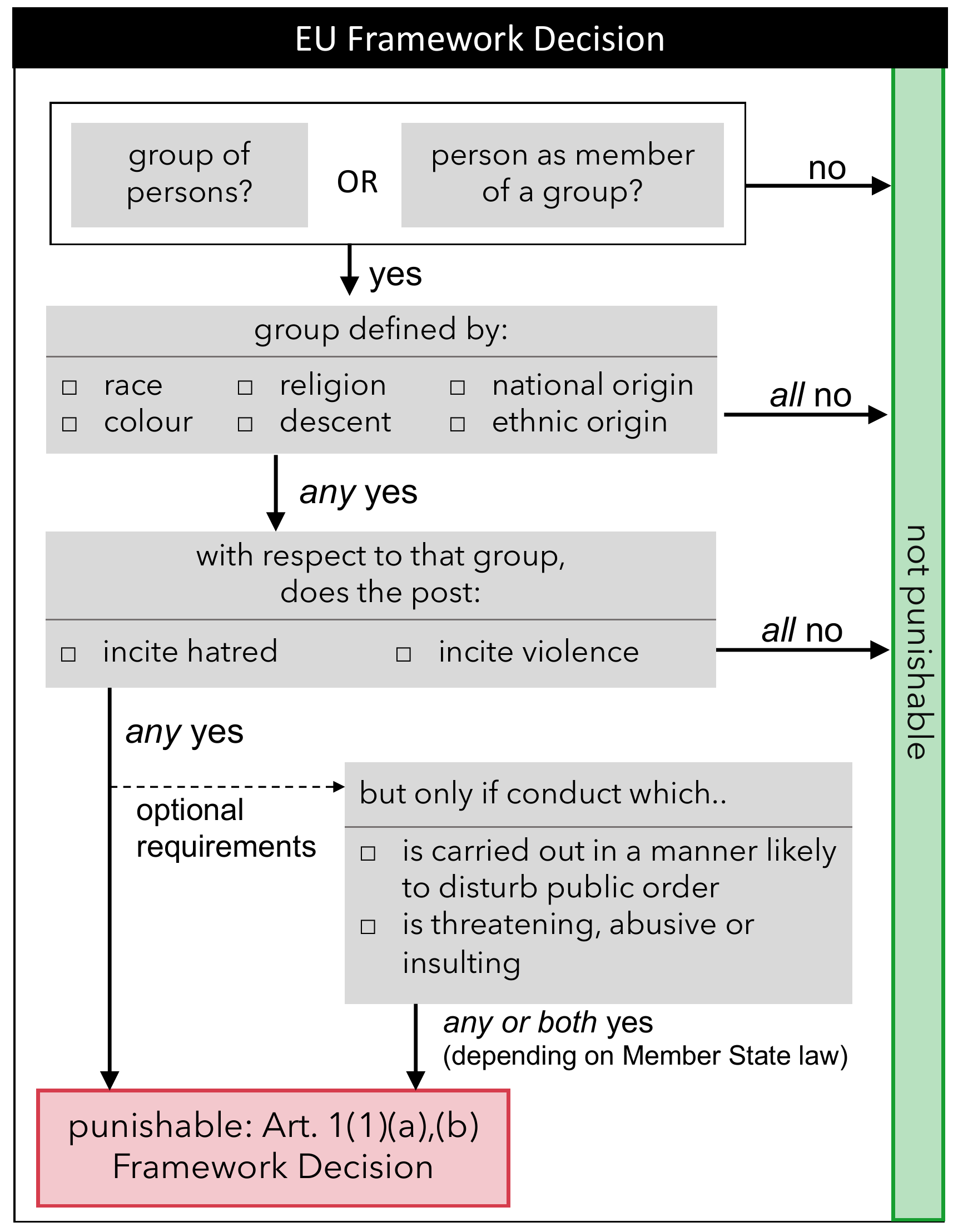}
\caption{Decision tree derived from legal framework}
\label{fig:decisiontree_EU}
\end{figure}

We begin our investigation by operationalizing the relevant part of the Framework Decision (FD) into a sequence of binary decisions that can be reliably annotated (see Figure~\ref{fig:decisiontree_EU} for the final decision tree).
In a way, we are translating the plain text of the legal definition into an actionable algorithm.

Article 1(1) FD states that the following intentional conduct is punishable:
\begin{quote}
\small
(a) publicly inciting to violence or hatred directed against a group of persons or a member of such a group defined by reference to race, colour, religion, descent or national or ethnic origin;

(b) the commission of an act referred to in point (a) by public dissemination or distribution of tracts, pictures or other material; 
\end{quote}


\noindent
The punishable conduct addressed in paragraph (a) refers to the oral expression of hatred, while paragraph (b) broadens the scope to public dissemination or distribution of tracts, pictures or other material.
For the detection of social-media posts, there is no added value in implementing these actions separately, as they are always met in case of public social-media posting on the Internet.



In a simplified way, two main questions have to be answered: (1) does a statement address a protected group? and (2) does it target that group by inciting hatred or violence?
We address these as (1) \textit{target group} and (2) \textit{targeting conduct}.

\subsection{Target Group}
\label{sec:target_group}

As shown in Figure~\ref{fig:decisiontree_EU}, Art.1(1)(a) refers to the following potential targets: a group of persons or a member of such a group defined by reference to race, colour, religion, descent, or national or ethnic origin (see Example~\ref{ex:distinguishable_groups}). 

\begin{example}
\scriptsize
\noindent\hspace*{2mm} \rule{200pt}{0.6pt}\\[-1,5ex]
\captionsetup{justification=raggedright, margin={10pt,5pt}}
\begin{enumerate}[nosep]
    \item[--] {\fontfamily{phv}\selectfont \scriptsize \textbf{French people} are frog eaters. (nationality)}
    \item[--] {\fontfamily{phv}\selectfont \scriptsize \textbf{Black people} = slaves!! (race)}
    \item[--] {\fontfamily{phv}\selectfont \scriptsize \textbf{Muslims} are all terrorists! (religion)}
    \item[--] {\fontfamily{phv}\selectfont \scriptsize \textbf{Sinti and Roma} - awful parasites! (ethnic origin)}
\end{enumerate}
\caption{Distinguishable groups.}
\label{ex:distinguishable_groups}
\noindent\hspace*{4mm} \rule{200pt}{0.6pt}\\[-4ex]
\end{example}
The scope also covers \textit{individuals} in case they are targeted as a member of an aforementioned group, as illustrated in Example~\ref{ex:individual_as_member}.

\begin{example}
\footnotesize
\noindent\hspace*{2mm} \rule{200pt}{0.6pt}\\[-2ex]
\captionsetup{justification=raggedright, margin={10pt,5pt}}
\begin{enumerate}[nosep]
    \item[--] {\fontfamily{phv}\selectfont \scriptsize you fucking muslim should leave our country!}
    \item[--] {\fontfamily{phv}\selectfont \scriptsize This dirty american bitch, typical american, lying son-of-a-bitch, out of our country!}
\end{enumerate}
\caption{Individuals as members of a group.}
\label{ex:individual_as_member}
\noindent\hspace*{2mm} \rule{200pt}{0.6pt}\\[-3ex]
\end{example}

`Race and `colour' are discriminating grounds that can be understood interchangeable.
`Religion' refers broadly to persons defined by reference to their religious convictions or beliefs (Recital (8)).


Recital (7) clarifies that `descent' points to persons or groups of persons who descend from persons who could be identified by characteristics like race or colour. 
It is not necessary that all these characteristics still be present in the respective persons.
Hence, the descendants would be protected regardless, e.g., descendants of people of Jewish faith even in cases where they do not share this faith anymore.
`National origin' or `ethnic origin' are both distinguishing grounds that require reference to a specific nationality or ethnic group. 
Statements that refer to `foreigners' or `refugees' without further specification are not covered, as these references are considered too general. 

\subsection{Targeting Conduct}
\label{sec:conduct}

With respect to the target group as a victim, Art.1(1)(a) requires at least one of the following acts to be committed by the potential offender: (i) inciting hatred, or (ii) inciting violence.

Regarding the definition and understanding of these acts, freedom of expression needs to be taken into consideration through Art.7(1), which ultimately refers to Art.11(1) of the EU Charter of Human Rights.
By preventing segregation, the intent is to protect minorities from being deprived of their human dignity as equal members of society.
Punishing expressions is only justified in the respective cases if the legal interest in preventing discrimination outweighs the right to free speech -- which is likewise a precondition for democracy.

Within these limits, the Framework Decision itself does not provide for a more detailed definition of `inciting hatred' and `inciting violence', but entrusts the Member States with elaborating the interpretation in national case law. 
For our annotation guidelines, we draw here from German case law, which provides for long-standing settled decision-making practice for these terms.

\paragraph{Inciting}
`Inciting' has been defined as `conduct influencing emotions and intellect of others'.\footnote{BGHSt 21, 371 (372); BGHSt 46, 212 (217)}
A key element of the definition is the clear intent to influence others.
To outweigh freedom of expression, the conduct has to go beyond mere rejection or contempt and means more than merely endorsing.

\paragraph{Hatred}
The Framework Decision limits, in Recital (8), the notion of `hatred' as such based on race, colour, religion, descent, or national or ethnic origin.
In other words, `hatred' expressed against a specific group, but which is unrelated to the belonging to this group, is not covered.
We draw here again on German case law, where the act of incitement to hatred needs to be `objectively capable and subjectively intended to create or intensify an emotionally enhanced, hostile attitude (towards the respective group)'.\footnote{BGHSt 21, 371 (372); BGHSt 46, 212 (217)}
Example~\ref{ex:inciting_hatred} illustrates comments that fit these criteria.


\begin{example}
\footnotesize
\noindent\hspace*{2mm} \rule{200pt}{0.6pt}\\[-2ex]
\captionsetup{justification=raggedright, margin={10pt,5pt}}
\begin{enumerate}[nosep, rightmargin=10pt]
    \item[--] {\fontfamily{phv}\selectfont \scriptsize Muslims are deceitful parasites enjoying life thanks to hard working german citizens!!
    }
     \item[--] {\fontfamily{phv}\selectfont \scriptsize Bring back the slaves! \#niggerarenohumans
    }
\end{enumerate}
\caption{Comments inciting hatred.}
\label{ex:inciting_hatred}
\noindent\hspace*{2mm} \rule{200pt}{0.6pt}\\[-6ex]
\end{example}

\paragraph{Violence}
While `hatred' refers to the creation of a hostile attitude, inciting `violence' shall `give rise to the determination of others to commit violence'.\footnote{BGH 3.4.2008 – 3 StR 394/07}
Violent measures do not just comprise assault, but also violent expulsion or pogroms. 
Example~\ref{ex:violent_measures} illustrates comments inciting violence.

\begin{example}
\footnotesize
\noindent\hspace*{2mm} \rule{200pt}{0.6pt}\\[-2ex]
\captionsetup{justification=raggedright, margin={10pt,5pt}}
\begin{enumerate}[nosep]
    \item[--] {\fontfamily{phv}\selectfont \scriptsize U.S. citizens should be hunt down and deported!}
    \item[--] {\fontfamily{phv}\selectfont \scriptsize Burn all Muslims in their mosques!}
\end{enumerate}
\caption{Comments inciting violence.}
\label{ex:violent_measures}
\noindent\hspace*{2mm} \rule{200pt}{0.6pt}\\[-6ex]
\end{example}


\subsection{Optional Qualifiers} 
\label{optional_requirements}
Art.1(2), however, grants one exception to the minimum standard, as seen in Figure~\ref{fig:EU_MS}. 
Member States may predicate the offense on the additional requirements of the disturbance of public order or threatening, abusive or insulting conduct.
In other words, a Member State may stipulate that the conduct is only punishable if it also leads to a disturbance of public order, or if the conduct is also threatening, abusive, or insulting.
As these additional requirements are only required by a few Member States, we do not operationalize them.


\section{Feasibility Study}
\label{sec:feasibility-study}

To test our decision tree annotation scheme, we first perform a feasibility study, where we assess the quality of annotations produced by our annotation scheme against direct annotation.
We also assess the reliability of an assessment by legal experts to establish an upper bound for this task.

\paragraph{Setup} We asked public prosecutors from one of the two cybercrime prosecution centers in Germany to provide the ground truth for punishability based on \S 130 of the German Criminal Code -- which implements the EU Framework.\footnote{As \S 130 of the German Criminal Code is a transposition of the minimum standard set by the EU Framework Decision (see Section~\ref{sec:operationalizing}), the results obtained in this way should be generalizable to EU law.}
As prosecutors would be obliged to open an investigation for each punishable post, we provided a set of 156 `made-up' hate speech posts in German.
These were never openly published and are thus not punishable.\footnote{The made-up posts are comparable in nature to realistic posts. See next Section~\ref{sec:dataset} for a more detailed description.}
The prosecutors did not use our decision tree, but decided based on their legal training and expertise.
As a control condition, we asked layperson annotators to perform a direct annotation.
Annotators were provided with the legal text of \S 130 and decided whether a post was punishable using their understanding of the legal code.
Finally, we asked layperson annotators to follow our multi-label annotation scheme, from which we can automatically derive whether a post is punishable or not, depending on the combination of our labels.

%

\begin{figure}
    \centering
    \includegraphics[width=.85\columnwidth]{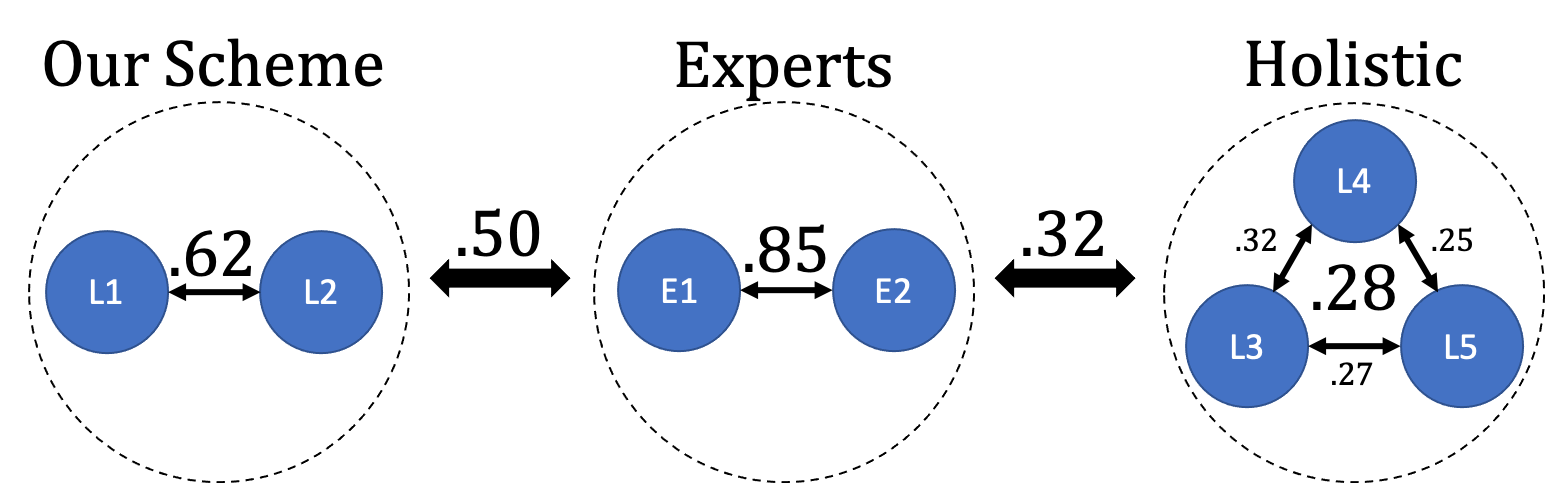}
    \caption{Cohen's Kappa for the \textit{Punishable} label for different annotation schemes in the feasibility study.}
    \label{fig:kappas}
\end{figure}

\paragraph{Results}
Figure~\ref{fig:kappas} shows the inter-annotator agreement (IAA) per setup in the feasibility study.
Agreement in the control condition (holistic annotation) is very low, which is in line with previous findings of low IAA for hate speech annotations \cite{Ross2016}.
However, the high kappa between expert prosecutors shows that sufficient legal expertise enables consistent judgements.

Using our annotation scheme increases consistency between annotators and agreement with experts.  
Thus, based on the success of the feasibility study, we adapt our annotation scheme to fit the EU framework and produced the full dataset, described in the next section.

\section{Punishable Hate Speech Dataset}
\label{sec:dataset}
In this section, we describe how the full dataset was created.
All posts in the dataset are in German.

\subsection{Data Sources}
Social-media posts were sampled and requested from a multitude of sources with the primary goal of obtaining sufficient examples of punishable hate speech.
Table~\ref{table:datasetsourcetable} provides an overview of the final composition of the dataset.

\begin{table}[t]
\footnotesize
\centering
\captionsetup{justification=centering}
\begin{tabular}{lrr} 
     \toprule
     Source  &  \# & \% Punishable   \\
     \midrule
     Made-up & 157 & 13.5 \\
     Web search & 80 & 6.2 \\
     Anti hate speech initiatives & 88 & 10.2   \\
     GermEval2019 (abuse, insult) & 425 & 0.9  \\
     GermEval2019 (other) & 250 & 0.0 \\ 
\bottomrule
\end{tabular}
     \captionof{table}{Composition of the dataset by source.}
     \label{table:datasetsourcetable}
\end{table}

\textbf{Made-up}
We include the `made-up' examples from the feasibility study, re-annotated according to the EU framework. 
The examples were produced by volunteers, who were instructed to write short texts presumably constituting `incitement to hatred' against the list of target groups mentioned in Figure~\ref{fig:confusion-matrix}.\footnote{Volunteers did not participate in subsequent annotation efforts.}
Participants also received instances of real hate speech as examples for their artificial posts.
9 participants created a total of 157 short texts.
The resulting statements are nearly indistinguishable in form from real examples, but we have no way of controlling for topic biases that might have been introduced via this process.


\textbf{Web search}
We performed a manual search of Twitter, comment sections of online newsrooms, law forums, court databases as well as news articles for hateful social media posts that were included in a court decision. This resulted in 80 instances.

\textbf{Anti hate speech initiatives}
We include 88 hate speech comments collected by the initiative `respect!' of Demokratiezentrum BW.


\textbf{GermEval2019}
Data samples from the subtask 2 corpus of GermEval 2019, a shared task on the identification of offensive language \cite{struss2019overview}, were also included.
We add 425 tweets from the `abuse' and `insult' categories, that will probably contain only few cases of legally punishable hate speech, but are likely to contain offensive language.
We also add 250 tweets from the `other' category that are non-offensive, but about similar topics (like refugee crisis, policits, etc.).

\subsection{Annotation Scheme \& Process}
\label{Annotation}

The full dataset was annotated by two paid laypersons, who were provided with an annotation manual based on the operationalisation described in Section~\ref{sec:operationalizing} with further explanations, instructions, and examples.
To measure annotation quality, a subset of 101 posts was annotated by a fully-qualified lawyer using the same annotation scheme.

We annotate whether a group of persons or a group member was mentioned in a post and, if so, whether the group is distinguishable by any reference to race, descent, or national or ethnic origin.
In case a group is explicitly mentioned, we also annotate the surface form used in the comment. 
We created a short list of frequently attacked groups and asked annotators to choose one of these or `other' (`Group Category' annotation).
We include groups not covered by the EU framework like \textit{women} or \textit{refugees}, as they might be relevant for future detection tasks regarding other legal regimes.
The full list of target groups used in our study can be seen in Figure~\ref{fig:confusion-matrix}.

        \begin{example}
        \scriptsize
        \noindent\hspace*{2mm} \rule{200pt}{0.6pt}\\[-2ex]
        \captionsetup{justification=raggedright, margin={10pt,5pt}}
        \begin{enumerate}[nosep, rightmargin=10pt]
             \item[--] {\fontfamily{phv}\selectfont \scriptsize Euthanasia is the only way to deal with all the \textbf{disabled people}, they should be buried.}
             \item[--] {\fontfamily{phv}\selectfont \scriptsize You should gas all the \textbf{Jews}. All they want is your money.}
             \item[--] {\fontfamily{phv}\selectfont \scriptsize The \textbf{oil eyes} should set themselves on fire with their gasoline.}
            \end{enumerate}
        \caption{Explicitly mentioned groups.}
        \label{ex:explicit_group}
        \noindent\hspace*{2mm} \rule{200pt}{0.6pt}\\[-2ex]
        \end{example}
        
        

We also annotate possible targeting conduct described in Section~\ref{sec:conduct}, i.e.\ inciting hatred and inciting violence.
If a relevant group is targeted and any targeting conduct is present, a post is considered punishable.
Table~\ref{tab:datasetExamples} provides some examples of the resulting annotation.
\newcommand*\rot{\rotatebox{90}}

\newcommand*\cm{\checkmark}

\begin{table*}[t]
\centering
\scriptsize
\captionsetup{justification=centering}
\begin{tabular}{@{}lll@{\hspace{1\tabcolsep}}l@{\hspace{1\tabcolsep}}lll@{\hspace{1\tabcolsep}}l@{}}
\toprule
            & & \multicolumn{3}{c}{\bf Target Group} & \multicolumn{2}{c}{\bf Targeting Conduct} &\\
\bf \makecell{Statement} & \bf \makecell{Group\\Category} & \rot{Group of} \rot{persons} & \rot{Distinguishable by} \rot{nationality, race,} \rot{religion, ethnicity}  & \multicolumn{1}{c|}{\rot{Explicit}} & {\rot{Incite hatred}} & \rot{Incite violence} & \rot{\bf Punishable} \\
\midrule
\makecell[l]{So dumm wie \underline{Gutmenschen} halt sind!\\Stupid as \underline{do-gooders} are!}  & Other & \cm &
               & \cm &   & & \\
\addlinespace[1mm]
\makecell[l]{\underline{Die linken} Hurens\"ohne! Ballert die alle ab!\\\underline{Left-wing} sons of bitches! Shoot 'em all!} & Left-Wing & \cm &
             & \cm &  & \cm   & \\
\addlinespace[1mm]
\makecell[l]{Es brennen noch zu wenige Synagogen.\\Still too few synagogues on fire.} & Jews & \cm & \cm 
         &&    &   \cm & \cm\\
\addlinespace[1mm]
\makecell[l]{\underline{Muslime} sind alles Vergewaltiger! Sch\"utzt deutsche Frauen!\\\underline{Muslims} are all rapists! Protect our German women!}
& Muslims & \cm &
             \cm &   \cm  & \cm &  &  \cm\\

\bottomrule
\end{tabular}
\caption{Example annotations from the resulting dataset. Surface form referring to groups is underlined.}
\label{tab:datasetExamples}
\end{table*}


\subsection{Analysis}

\begin{table}
\footnotesize
\centering
\begin{tabular}{cllll} 
\toprule
&   & L1/ & L1/ & L2/ \\ 
&   & L2 & Exp & Exp \\ 
\midrule
& \bf Group Category                                                                        & \bf .77      & \bf .70   & \bf .67                                                                             \\ 
\addlinespace[2mm]
\multirow{4}{*}{\rot{\bf{\makecell{Group}}}} 
& Group of persons                                                                     & .49   & .82   & .55                                                                             \\ 

& Individual as group member                                                            & .14   & .24   & .48                                                                            \\
& Nationality, race, religion, ...
        & .52    & .42    & .67                                                                             \\
& \textbf{Any target group}                                                         & \bf .53    & \bf .52    & \bf .70                                                                             \\ 
\addlinespace[2mm]
\multirow{3}{*}{\rot{\bf{Conduct}}}  

& Inciting hatred                                                                      & .11     & .39    & .00                                                                             \\
 & Inciting violence                                                              & .56     & .64      & .74                                                                             \\ 
  & \textbf{Any targeting conduct}                                                 & \bf .56     & \bf .69     & \bf .52                                                \\ 
\addlinespace[2mm]

& \textbf{Punishable}            
    & \bf .33   & \bf .43   & \bf .37                                                        \\
\bottomrule
\end{tabular}

\caption{Inter-annotator agreement (Cohen's Kappa) between laypersons and domain expert.}
\label{tab:kappa}
\end{table}

We analyze the IAA among laypersons as well as between laypersons and the expert annotator in terms of \emph{Cohen's Kappa} as shown in Table~\ref{tab:kappa}.
Aggregated results on target group and targeting conduct are quite reliable (kappa between .52 and .70), while kappa for the punishable label is rather low (.33 to .43).
People agree on the facts (group, conduct), but disagree on the interpretation.

Table~\ref{table:datasetsourcetable} displays the distribution of punishable posts.
It is noteworthy that in the GermEval2019 data a surprisingly low proportion (under 1\%) of abusive or insulting comments are actually punishable under EU law.
This highlights that hate speech detection and detecting illegal content are fundamentally different tasks.

\begin{figure}[t]
    \scriptsize
    \centering
    \includegraphics[scale=0.5]{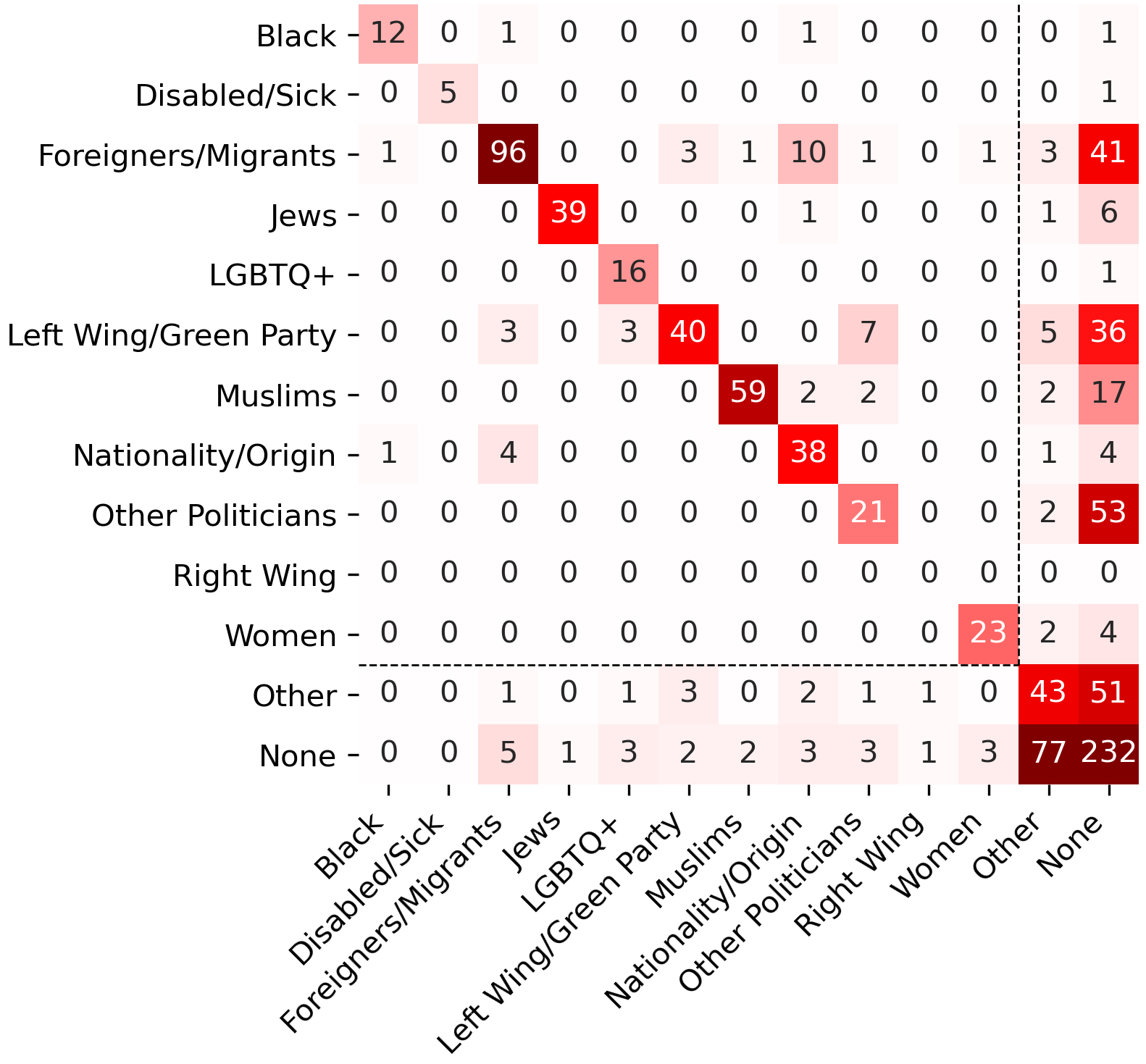}
    \caption{Confusion matrix of non-expert annotators.}
    \label{fig:confusion-matrix}
\end{figure}

Figure~\ref{fig:confusion-matrix} shows the confusion matrix between the two layperson annotators regarding the group annotation from our short list (subset of 392 posts having a group mention).
The largest target group is foreigners/migrants, which is not explicitly protected under EU law.
Differences between annotators mainly arise due to the `None' and `Other' categories, while the largest disagreement is within closely related categories like `left-wing/green party' and `other politicians'.

Each group is referred to by a wide variety of different surface forms.
Table~\ref{tab:surface-forms} lists selected examples of surface forms in the dataset.
The median number of surface forms per group is 20 (min=3, max=135), showing that automatic detection will have to deal with a high variance.
The `other' category contains a wide range of different types of groups like law enforcement, vegans, jobless, football clubs, or media outlets that we might consider as distinct groups in a revised annotation scheme.


\section{Automated Detection}

To study the extent to which our annotated data can serve as a basis for automated detection, we train a baseline classifier that takes a post as input and estimates whether the post is punishable. For model training, differences between annotators were adjudicated by a legal expert. The agreement numbers reported in Section~\ref{sec:dataset} are thus not applicable for the following experiment.

\paragraph{Setup}
Fine-tuned BERT~\citep{Devlin2018} models have proven to be strong baselines for various NLP tasks, so we follow this practice\footnote{For the implementation, we use \textit{HuggingFace Transformers}~\citep{Wolf2020} and \textit{PyTorch}~\citep{Paszke2019}.},
using GBERT-base~\citep{Chan2020}.
The model is trained for 20 epochs using a batch size of 16 and NLL loss.
For optimization, we choose bias-corrected Adam \citep{Kingma2015adam}, with a learning rate of $2e^{-5}$.
The learning rate is linearly increased up to its peak during the first 10\% of training and then linearly decreased.
These choices follow the recommendations of \citet{Mosbach2020} for increasing training stability when fine-tuning BERT.
For evaluation, we perform a stratified 10-fold CV.
We also include a random classification baseline for the punishable label.
We set $p=0.5$ and thus yield a recall of 0.5 and a precision that corresponds to the overall ratio of punishable posts in the dataset, i.e.\ 0.04.

\begin{table}[t]
\centering
\footnotesize
\begin{tabular}{@{}lccc@{}}
\toprule
                                      & P       & R    & $F_1$\\ \midrule
Group of persons                      & .81     & .85  & .83 \\
Individual as member of group         & .00     & .00  & .00 \\
Distinguishable by nationality, etc.  & .79     & .71  & .75 \\ 
\addlinespace[1.5mm]
Inciting hatred                       & .25     & .07  & .11 \\
Inciting violence                     & .70     & .73  & .72 \\ 
\addlinespace[1.5mm]
Punishable (random)                   & .04     & .50  & .07 \\
Punishable (direct)                   & .69     & .28  & .39 \\
Punishable (submodels + decision tree)& .41     & .43  & .42 \\ \bottomrule
\end{tabular}

\caption{Overview of prediction results. See Table~\ref{tab:dataset-dist} in the Appendix for label distributions.}
\label{tab:performance}
\end{table}


\paragraph{Results}
The fine tuned model achieves an average $F_1$ of .39 (P .69; R .28), which shows that legal hate speech classification is complex and not easily solvable by baseline models.
The mere presence of language inciting hate or violence is not a sufficient signal, but the model needs to learn in addition whether (i) the hate is directed against an object, (ii) the object is a group, (iii) the group is protected under the given law.



This is exemplified looking at some misclassifications.
In Example~\ref{ex:hard-to-learn}.1, no group is targeted; yet, the model classified the post as punishable. 
In Example~\ref{ex:hard-to-learn}.2, the model erroneously judged the post to be punishable, even though the group is not protected under the EU framework. 
Implicit or metaphorical references to a group were not detected by the model (e.g.\  Example~\ref{ex:hard-to-learn}.3).

\begin{example}
\scriptsize
\noindent\hspace*{2mm} \rule{200pt}{0.6pt}\\[-2ex]
\captionsetup{justification=raggedright, margin={10pt,5pt}}
\begin{enumerate}[nosep, rightmargin=10pt]
     \item[1)] {\fontfamily{phv}\selectfont \scriptsize DEPORT DEPORT [...] DEPORT}
     \item[2)] {\fontfamily{phv}\selectfont \scriptsize Faggots should be castrated and locked up!}
     \item[3)] {\fontfamily{phv}\selectfont \scriptsize A metro we build, a metro we build, a metro we build from Jerusalem to Auschwitz, a metro we build!}
    \end{enumerate}
\caption{Cases misclassified by detection model}
\label{ex:hard-to-learn}
\noindent\hspace*{2mm} \rule{200pt}{0.6pt}\\[-2ex]
\end{example}



We also trained separate models for the prediction of target groups and targeting conduct. The last row in Table~\ref{tab:performance} shows results of applying those two models to derive punishability. In terms of $F_1$ score, the subtask approach is comparable to the direct approach of estimating punishability (.42 vs .39).
Looking at the performance of the subtask models, we see mixed results.
While the \textit{Group of persons}, \textit{Distinguishable by nationality, race, religion, ethnicity} and \textit{Inciting violence} models produce good results ($.71 - .83$), the models for \textit{Individual as member of group} and \textit{Inciting hatred} failed making accurate predictions ($.00 - .11$).
Both are rare in the dataset (14 positive cases each), making them difficult labels to learn.

\section{Generalizing beyond EU Law}

So far, we have presented a case study of operationalizing a specific legal standard (i.e.\ the EU Framework Decision).
The underlying methodology can be generalized in a straightforward way.
Instead of directly predicting whether a post is punishable or not, we divide the problem into two subtasks, (i) group detection and (ii) conduct detection, each of which can be tackled separately, depending on the applicable legal regime.
This approach offers higher explainability of model decisions, an aspect that is crucial for legal decision-making. 

\subsection{Group Detection}

If we were able reliably to detect all groups referred to in a comment, we could take the list of protected groups and only consider those relevant under a certain legal standard.
In this way, our approach would also generalize beyond EU law.

\begin{table}[]
\scriptsize
\begin{tabular}{p{1cm}p{6cm}}
\toprule
\textbf{Category}                            & \textbf{Surface Form}\\ \midrule
People of Color                  & \#negersindkeinemenschen, affe, bimbo, dunkler teint, nafris, neger, negroide goldst{\"u}cke, schwarze, sklaven\\
Jews                             & dreckiges judenpack, judenschwein, zentralrat der juden, j{\"u}discher zombie, rattenvolk, zionisten \\
Muslims                           & \#islamisierung, \#muslime, islamlobbys, b{\"a}rtigen kindersch{\"a}nder, ditib imams, dreckige kopftuchm{\"a}dchen, gotteskrieger, isis-schlampen, muslim-ungeziefer, scharia\\
Nationality/ Origin                            & pro-erdogan t{\"u}rken, abschaum afrikas, araber, schlitz{\"a}ugige, deutsche kartoffel, deniz, nafris, polnische hurens{\"o}hne\\
\bottomrule
\end{tabular}
\caption{Examples of surface forms of target groups}
\label{tab:surface-forms}
\end{table}

However, groups are often referenced by a variety of different surface forms, some of which are only metaphorically related to the group (e.g.\ `Goldst{\"u}cke'; Engl.\ `\textit{gold pieces}' for \textit{people of color}, see Table~\ref{tab:surface-forms}). 
Consequently, we cannot use Named Entity Recognition \cite{ritter2011named} for group detection, as, e.g.\ `women' are a common target group, but not a named entity.
A better fit seems Entity Linking \cite{derczynski2015analysis}, which would (depending on the underlying knowledge base) find explicitly mentioned groups.
However, groups can also be implicitly mentioned (\ref{ex:implicit}.1) 
or as part of a co-reference chain (\ref{ex:implicit}.2).

\begin{example}
\scriptsize
\noindent\hspace*{2mm} \rule{200pt}{0.6pt}\\[-2ex]
\captionsetup{justification=raggedright, margin={10pt,5pt}}
\begin{enumerate}[nosep, rightmargin=10pt]
     \item[1)] {\fontfamily{phv}\selectfont \scriptsize [...] For them the sport [football] is like. I put a goat on the field, 22 holy warriors and whoever knocks it up first, wins.}
     \item[2)] {\fontfamily{phv}\selectfont \scriptsize No mercy for \textbf{terrorists}. We have declared war on \textbf{Islam}. \textbf{They} had 800 years to reform. Time is up!}
    \end{enumerate}
\caption{1) Implicit targeting of Muslims. 2) Muslims target group only identifiable by coreference.}
\label{ex:implicit}
\noindent\hspace*{2mm} \rule{200pt}{0.6pt}\\[-2ex]
\end{example}

Thus, we argue that annotating data for groups referenced in the text (even implicitly) is a prerequisite for `group detection' as a stand-alone NLP task.
Once this is established, it can be used to find the best methods for group detection.
A possible way to find surface variants might be to compile a list of common surface forms and compare the closest synonyms for a group as computed over a more general corpus.


\subsection{Conduct Detection}
For specific targeting conduct like \textit{inciting violence}, detecting the most common actions patterns like `kill \textsc{Group}' or `burn \textsc{Group}' might be a promising approach. 
This would also limit the number of false positives, e.g.\ when someone `threatens' to \textit{burn a candle} instead.
For this task, semantic role labeling \cite{Gildea2000SRL} or using frames \cite{Baker2014FRAMES} could be useful, but existing resources like FrameNet seem not specific enough, as they put `threat' under the \textsc{Commitment} frame (in the sense of `committing to harm someone').


In general, there is a high level of metaphor, irony and sarcasm in the comments, which poses serious challenges to all conduct detection methods.
Even though irony and sarcasm are not legal terms as such, they might have an influence on the assessment as to whether a targeting conduct like \textit{inciting hatred} is given.
Accordingly, these cases can be captured at the annotation level as \textit{in dubio pro reo}, i.e.\ not punishable.

\section{Related Work}
Automated detection of offensive Internet discourse has been intensively studied under a variety of names, for instance: abusive language \cite{Waseem2017} or content \cite{Kiritchenko2020}, ad hominem arguments \cite{Habernal2018}, aggression \cite{Kumar2018}, cyberbullying \cite{Xu2012,Macbeth2013}, hate speech \cite{Warner2012,Ross2016,DelVigna2017}, offensive language usage \cite{Razavi2010}, profanity \cite{Schmidt2017}, threats \cite{Oostdijk2013} and socially unacceptable discourse \cite{Fiser2017}.
While most early work focused on English, now there is also a growing body of work in other languages, e.g., German \cite{Ross2016}, Italian \cite{DelVigna2017}, and Dutch \cite{Oostdijk2013}.
All of those works use a non-legally informed definition of the construct to be detected.
A notable exception is work on Slovene by \citet{Fiser2017} who relied on annotators interpretation of the legal definition without further breaking down those decisions.

There is a body of work in NLP with a legal perspective focused on predicting the outcome of court decisions \cite{Aletras2016, Katz2017, Bruninghaus2003, Kastellec2010, Waltl2017}, to the best of our knowledge our work is the only effort to operationalise a legal framework for hate speech.
However, the dependence on existing court decisions makes it difficult to work with legal problems where relevant case law is not available as a data source. To overcome this problem, \citet{Zufall2019} translated statutory rules for defamatory offenses into a series of annotatable binary decisions.

The importance of finding groups for hate speech analysis has also been stressed by \newcite{Kiritchenko2020}.
As offenses against groups are often implicitly framed, \newcite{Sap2020} introduce \textit{Social Bias Frames} that make the attacked group explicit.
As group detection can work with any set of group categories, it can also be adapted to cover non-Western groups \cite{sambasivan2021reimagining}.

\section{Conclusion}
We operationalize a `legal approach to hate speech' by translating the requirements of the EU Framework Decision into a series of annotation steps that can be reliably performed by laypersons.
However, we show that learning a model of whether a post is punishable remains challenging.
We thus propose to tackle two subtasks instead: \textit{group detection} and \textit{conduct detection}.
Depending on the applicable legal framework, a final decision on the legal status of a comment can then be derived from the combination of detected group and conduct.
Relying on subtasks comes with the added benefit of increased transparency and explainability compared to black-box models.
This is crucial for systems that potentially interfere with human rights, such as the balance between freedom of expression and the prevention of discrimination. 
Hence, we recommend this modular approach as the preferred way of composing systems for legal decision-making.

\newpage

\section*{Ethical Considerations}
Predicting the legal status of a comment might infringe on the fundamental right of `free speech'.
On the other hand, we are targeting the worst tail-end of the distribution -- the kind of hate speech that is putting democracy in danger by inciting hatred and violence in a society.
Not addressing hate speech and its foregoing automated detection methods would give further rise to possible discrimination, making it a problem for equal participation in a democracy.
As our approach introduces a layer of algorithmic transparency not found in traditional methods, we believe that the importance of this research outweighs its dangers.

\paragraph{Annotation Process}
Regarding our made-up examples, we conducted a survey with nine students, asking them to create short texts that presumably constitute `incitement to hatred' (see Section~\ref{sec:dataset}).
This survey was approved by the ethics committee of \textsc{Anonymized}.
The final annotation of the dataset was carried out by two paid annotators, who were compensated above the local minimum wage.
Annotators were warned about the offensive nature of the data and instructed only to annotate 50 comments a day to mitigate the effect of fatigue.

\paragraph{Race and Gender}
The EU Framework Decision explicitly requires the conduct to be directed against a ``group of persons or a member of such a group defined by reference to race, colour, religion, descent or national or ethnic origin'' (Art.1(1)(a) Framework Decision). 
It is thus a necessary legal requirement which is meant to protect the aforementioned groups and to prevent discrimination.
We also use the groups `women' and `LGBTQ+', as these are often the targets of hate speech.
Our model explicitly allows for adding other groups in order to adapt to differing legal standards.

\paragraph{Deploying Systems for Legal Decision-making}
Systems used in the context of legal decision-making or, more generally, systems that filter specific content should be used with great care and in view of the potential interference with human rights, namely the right to free speech. 
We explicitly do not recommend using any legal decision-making system without human supervision.
We consider the improved transparency of our model to be an important step in allowing prosecutors to understand the reasons behind flagging a certain comment as potentially punishable.

\paragraph{Release of the Data}
As our dataset consists of postings that could be traced back to individuals, it contains personal data in the sense of the EU General Data Protection Regulation (GDPR). 
To comply with this legal standard, and given the sensitive nature of the data, we do not make any of the real postings publicly available. We do, however, publish the made-up examples generated during the feasibility study.

\newpage

\appendix

\section{Data Distribution in Automated Experiments}
\label{sec:A}

Note that the number of total annotations is larger than 1000, since some posts contained multiple groups.

\begin{table}[!h]
\scriptsize
\centering
\begin{tabular}{@{}llr@{}}
\toprule
Annotation                           & false & true \\ \midrule
Group of persons                     & 541   & 465  \\
Individual as member of group        & 992   & 14   \\
Distinguishable by nationality, etc. & 744   & 262  \\
\addlinespace[1.5mm]
Inciting hatred                      & 992   & 14   \\
Inciting violence                    & 886   & 120  \\ 
\addlinespace[1.5mm]
Punishable                           & 966   & 40   \\ \bottomrule
\end{tabular}
\caption{Label distribution for automated detection experiments.}
\label{tab:dataset-dist}
\end{table}

\begin{table}[!h]
\scriptsize
\centering
\begin{tabular}{@{}lr@{}}
\toprule
                      & Group Category \\ \midrule
None                  & 341            \\
Foreigners/Migrants   & 155            \\
Other                 & 103            \\
Left Wing/Green Party & 93             \\
Muslims               & 81             \\
Other Politicians     & 69             \\
Nationality/Origin    & 49             \\
Jews                  & 46             \\
Women                 & 29             \\
LGBTQ+                & 17             \\
People of Color       & 15             \\
Disabled/Sick         & 6              \\
Right Wing            & 0              \\ \bottomrule
\end{tabular}
\caption{Distribution of adjudicated group categories in the dataset.}
\label{tab:group-distribution}
\end{table}

\bibliography{anthology,custom}

\begin{thebibliography}{36}
\expandafter\ifx\csname natexlab\endcsname\relax\def\natexlab#1{#1}\fi

\bibitem[{Aletras et~al.(2016)Aletras, Tsarapatsanis, Preoţiuc-Pietro, and
  Lampos}]{Aletras2016}
Nikolaos Aletras, Dimitrios Tsarapatsanis, Daniel Preoţiuc-Pietro, and
  Vasileios Lampos. 2016.
\newblock \href {https://doi.org/10.7717/peerj-cs.93} {{Predicting judicial
  decisions of the European Court of Human Rights: A Natural Language
  Processing perspective}}.
\newblock \emph{PeerJ Computer Science}.

\bibitem[{Baker(2014)}]{Baker2014FRAMES}
Collin~F. Baker. 2014.
\newblock \href {https://doi.org/10.3115/v1/W14-3001} {{F}rame{N}et: A
  knowledge base for natural language processing}.
\newblock In \emph{Proceedings of Frame Semantics in {NLP}: A Workshop in Honor
  of Chuck {F}illmore (1929-2014)}, pages 1--5, Baltimore, MD, USA. Association
  for Computational Linguistics.

\bibitem[{Bruninghaus and Ashley(2003)}]{Bruninghaus2003}
Stefanie Bruninghaus and Kevin~D. Ashley. 2003.
\newblock {Predicting Outcomes of Case Based Legal Arguments}.
\newblock In \emph{Proceedings of the International Conference on Artificial
  Intelligence and Law}, pages 233--242, New York, NY, USA. ACM.

\bibitem[{Chan et~al.(2020)Chan, Schweter, and M{\"{o}}ller}]{Chan2020}
Branden Chan, Stefan Schweter, and Timo M{\"{o}}ller. 2020.
\newblock \href {https://www.aclweb.org/anthology/2020.coling-main.598}
  {{German's Next Language Model}}.
\newblock In \emph{Proceedings of the 28th International Conference on
  Computational Linguistics}, pages 6788--6796, Barcelona, Spain (Online).
  International Committee on Computational Linguistics.

\bibitem[{Del~Vigna et~al.(2017)Del~Vigna, Cimino, Dell'Orletta, Petrocchi, and
  Tesconi}]{DelVigna2017}
Fabio Del~Vigna, Andrea Cimino, Felice Dell'Orletta, Marinella Petrocchi, and
  Maurizio Tesconi. 2017.
\newblock {Hate Me, Hate Me Not: Hate Speech Detection on Facebook}.
\newblock In \emph{Proceedings of the First Italian Conference on Cybersecurity
  (ITASEC17)}, pages 86--95.

\bibitem[{Derczynski et~al.(2015)Derczynski, Maynard, Rizzo, Van~Erp, Gorrell,
  Troncy, Petrak, and Bontcheva}]{derczynski2015analysis}
Leon Derczynski, Diana Maynard, Giuseppe Rizzo, Marieke Van~Erp, Genevieve
  Gorrell, Rapha{\"e}l Troncy, Johann Petrak, and Kalina Bontcheva. 2015.
\newblock Analysis of named entity recognition and linking for tweets.
\newblock \emph{Information Processing \& Management}, 51(2):32--49.

\bibitem[{Devlin et~al.(2018)Devlin, Chang, Lee, and Toutanova}]{Devlin2018}
Jacob Devlin, Ming{-}Wei Chang, Kenton Lee, and Kristina Toutanova. 2018.
\newblock \href {http://arxiv.org/abs/1810.04805} {{BERT:} pre-training of deep
  bidirectional transformers for language understanding}.
\newblock \emph{CoRR}, abs/1810.04805.

\bibitem[{Fisch(2002)}]{Fisch2002}
William~B. Fisch. 2002.
\newblock Hate speech in the constitutional law of the united states.
\newblock \emph{The American Journal of Comparative Law}, (50):463–492.

\bibitem[{Fi{\v{s}}er et~al.(2017)Fi{\v{s}}er, Erjavec, and
  Ljube{\v{s}}i{\'{c}}}]{Fiser2017}
Darja Fi{\v{s}}er, Toma{\v{z}} Erjavec, and Nikola Ljube{\v{s}}i{\'{c}}. 2017.
\newblock Legal framework, dataset and annotation schema for socially
  unacceptable online discourse practices in slovene.
\newblock In \emph{Proceedings of the First Workshop on Abusive Language
  Online}, pages 46--51. Association for Computational Linguistics.

\bibitem[{Fortuna and Nunes(2018)}]{Fortuna2018}
Paula Fortuna and S\'{e}rgio Nunes. 2018.
\newblock \href {https://doi.org/10.1145/3232676} {A survey on automatic
  detection of hate speech in text}.
\newblock 51(4).

\bibitem[{Fortuna et~al.(2021)Fortuna, Soler-Company, and Wanner}]{Fortuna2021}
Paula Fortuna, Juan Soler-Company, and Leo Wanner. 2021.
\newblock \href {https://doi.org/https://doi.org/10.1016/j.ipm.2021.102524}
  {{How well do hate speech, toxicity, abusive and offensive language
  classification models generalize across datasets?}}
\newblock \emph{Information Processing {\&} Management}, 58(3):102524.

\bibitem[{Gildea and Jurafsky(2002)}]{Gildea2000SRL}
Daniel Gildea and Daniel Jurafsky. 2002.
\newblock \href {https://doi.org/10.1162/089120102760275983} {Automatic
  labeling of semantic roles}.
\newblock \emph{Comput. Linguistics}, 28(3):245--288.

\bibitem[{Habernal et~al.(2018)Habernal, Wachsmuth, Gurevych, and
  Stein}]{Habernal2018}
Ivan Habernal, Henning Wachsmuth, Iryna Gurevych, and Benno Stein. 2018.
\newblock {Before Name-Calling: Dynamics and Triggers of Ad Hominem Fallacies
  in Web Argumentation}.
\newblock In \emph{Proceedings of the 2018 Conference of the North American
  Chapter of the Association for Computational Linguistics: Human Language
  Technologies, Volume 1 (Long Papers)}, pages 386--396. Association for
  Computational Linguistics.

\bibitem[{Kastellec(2010)}]{Kastellec2010}
Jonathan Kastellec. 2010.
\newblock {The Statistical Analysis of Judicial Decisions and Legal Rules with
  Classification Trees}.
\newblock \emph{Journal of Empirical Legal Studies}, 7(2):202--230.

\bibitem[{Katz et~al.(2017)Katz, Bommarito, and Blackman}]{Katz2017}
Daniel~Martin Katz, Michael~J. Bommarito, II, and Josh Blackman. 2017.
\newblock {A general approach for predicting the behavior of the Supreme Court
  of the United States}.
\newblock \emph{PLOS ONE}, 12(4):1--18.

\bibitem[{Kingma and Ba(2015)}]{Kingma2015adam}
Diederik~P. Kingma and Jimmy Ba. 2015.
\newblock \href {http://arxiv.org/abs/1412.6980} {Adam: {A} method for
  stochastic optimization}.
\newblock In \emph{3rd International Conference on Learning Representations,
  {ICLR} 2015, San Diego, CA, USA, May 7-9, 2015, Conference Track
  Proceedings}.

\bibitem[{Kiritchenko et~al.(2020)Kiritchenko, Nejadgholi, and
  Fraser}]{Kiritchenko2020}
Svetlana Kiritchenko, Isar Nejadgholi, and Kathleen~C. Fraser. 2020.
\newblock \href {http://arxiv.org/abs/2012.12305} {Confronting abusive language
  online: A survey from the ethical and human rights perspective}.

\bibitem[{Kumar et~al.(2018)Kumar, Ojha, Malmasi, and Zampieri}]{Kumar2018}
Ritesh Kumar, Atul~Kr. Ojha, Shervin Malmasi, and Marcos Zampieri. 2018.
\newblock {Benchmarking Aggression Identification in Social Media}.
\newblock In \emph{{Proceedings of the First Workshop on Trolling, Aggression
  and Cyberbullying (TRAC-2018)}}, pages 1--11. Association for Computational
  Linguistics.

\bibitem[{Macbeth et~al.(2013)Macbeth, Adeyema, Lieberman, and
  Fry}]{Macbeth2013}
Jamie Macbeth, Hanna Adeyema, Henry Lieberman, and Christopher Fry. 2013.
\newblock {Script-based story matching for cyberbullying prevention}.
\newblock In \emph{ACM SIGCHI Conference on Human Factors in Computing
  Systems}, pages 901--906.

\bibitem[{Mosbach et~al.(2020)Mosbach, Andriushchenko, and
  Klakow}]{Mosbach2020}
Marius Mosbach, Maksym Andriushchenko, and Dietrich Klakow. 2020.
\newblock \href {http://arxiv.org/abs/2006.04884} {{On the Stability of
  Fine-tuning BERT: Misconceptions, Explanations, and Strong Baselines}}.
\newblock \emph{arXiv}.

\bibitem[{Oostdijk and van Halteren(2013)}]{Oostdijk2013}
Nelleke Oostdijk and Hans van Halteren. 2013.
\newblock {N-Gram-Based Recognition of Threatening Tweets}.
\newblock In \emph{Computational Linguistics and Intelligent Text Processing},
  pages 183--196, Berlin, Heidelberg. Springer Berlin Heidelberg.

\bibitem[{Paszke et~al.(2019)Paszke, Gross, Massa, Lerer, Bradbury, Chanan,
  Killeen, Lin, Gimelshein, Antiga, Desmaison, Kopf, Yang, DeVito, Raison,
  Tejani, Chilamkurthy, Steiner, Fang, Bai, and Chintala}]{Paszke2019}
Adam Paszke, Sam Gross, Francisco Massa, Adam Lerer, James Bradbury, Gregory
  Chanan, Trevor Killeen, Zeming Lin, Natalia Gimelshein, Luca Antiga, Alban
  Desmaison, Andreas Kopf, Edward Yang, Zachary DeVito, Martin Raison, Alykhan
  Tejani, Sasank Chilamkurthy, Benoit Steiner, Lu~Fang, Junjie Bai, and Soumith
  Chintala. 2019.
\newblock \href
  {http://papers.neurips.cc/paper/9015-pytorch-an-imperative-style-high-performance-deep-learning-library.pdf}
  {Pytorch: An imperative style, high-performance deep learning library}.
\newblock In H.~Wallach, H.~Larochelle, A.~Beygelzimer, F.~d\' Alch\'{e}-Buc,
  E.~Fox, and R.~Garnett, editors, \emph{Advances in Neural Information
  Processing Systems 32}, pages 8024--8035. Curran Associates, Inc.

\bibitem[{Poletto et~al.(2020)Poletto, Basile, Sanguinetti, Bosco, and
  Patti}]{Poletto2020}
Fabio Poletto, Valerio Basile, Manuela Sanguinetti, Cristina Bosco, and Viviana
  Patti. 2020.
\newblock \href {https://doi.org/10.1007/s10579-020-09502-8} {{Resources and
  benchmark corpora for hate speech detection: a systematic review}}.
\newblock \emph{Language Resources and Evaluation}, pages 1--47.

\bibitem[{Razavi et~al.(2010)Razavi, Inkpen, Uritsky, and Matwin}]{Razavi2010}
Amir~H. Razavi, Diana Inkpen, Sasha Uritsky, and Stan Matwin. 2010.
\newblock Offensive language detection using multi-level classification.
\newblock In \emph{{Proceedings of the 23rd Canadian Conference on Advances in
  Artificial Intelligence}}, pages 16--27, Berlin, Heidelberg. Springer-Verlag.

\bibitem[{Ritter et~al.(2011)Ritter, Clark, Etzioni et~al.}]{ritter2011named}
Alan Ritter, Sam Clark, Oren Etzioni, et~al. 2011.
\newblock Named entity recognition in tweets: an experimental study.
\newblock In \emph{Proceedings of the 2011 conference on empirical methods in
  natural language processing}, pages 1524--1534.

\bibitem[{Ross et~al.(2016)Ross, Rist, Carbonell, Cabrera, Kurowsky, and
  Wojatzki}]{Ross2016}
Björn Ross, Michael Rist, Guillermo Carbonell, Ben Cabrera, Nils Kurowsky, and
  Michael Wojatzki. 2016.
\newblock {Measuring the Reliability of Hate Speech Annotations: The Case of
  the European Refugee Crisis}.
\newblock In \emph{{Proceedings of NLP4CMC III: 3rd Workshop on Natural
  Language Processing for Computer-Mediated Communication}}, pages 6--9.

\bibitem[{Sambasivan et~al.(2021)Sambasivan, Arnesen, Hutchinson, Doshi, and
  Prabhakaran}]{sambasivan2021reimagining}
Nithya Sambasivan, Erin Arnesen, Ben Hutchinson, Tulsee Doshi, and Vinodkumar
  Prabhakaran. 2021.
\newblock \href {http://arxiv.org/abs/2101.09995} {Re-imagining algorithmic
  fairness in india and beyond}.

\bibitem[{Sap et~al.(2020)Sap, Gabriel, Qin, Jurafsky, Smith, and
  Choi}]{Sap2020}
Maarten Sap, Saadia Gabriel, Lianhui Qin, Dan Jurafsky, Noah~A Smith, and Yejin
  Choi. 2020.
\newblock \href {https://doi.org/10.18653/v1/2020.acl-main.486} {{Social Bias
  Frames: Reasoning about Social and Power Implications of Language}}.
\newblock In \emph{Proceedings of the 58th Annual Meeting of the Association
  for Computational Linguistics}, pages 5477--5490, Stroudsburg, PA, USA.
  Association for Computational Linguistics.

\bibitem[{Schmidt and Wiegand(2017)}]{Schmidt2017}
Anna Schmidt and Michael Wiegand. 2017.
\newblock {A Survey on Hate Speech Detection using Natural Language
  Processing}.
\newblock In \emph{Proceedings of the Fifth International Workshop on Natural
  Language Processing for Social Media}, pages 1--10. Association for
  Computational Linguistics.

\bibitem[{Stru{\ss} et~al.(2019)Stru{\ss}, Siegel, Ruppenhofer, Wiegand, and
  Klenner}]{struss2019overview}
Julia~Maria Stru{\ss}, Melanie Siegel, Josep Ruppenhofer, Michael Wiegand, and
  Manfred Klenner. 2019.
\newblock Overview of germeval task 2, 2019 shared task on the identification
  of offensive language.
\newblock In \emph{Proceedings of the 15th Conference on Natural Language
  Processing (KONVENS 2019)}, pages 354--365, Erlangen, Germany. German Society
  for Computational Linguistics \& Language Technology.

\bibitem[{Waltl et~al.(2017)Waltl, Bonczek, Scepankova, Landthaler, and
  Matthes}]{Waltl2017}
Bernhard Waltl, Georg Bonczek, Elena Scepankova, J{\"o}rg Landthaler, and
  Florian Matthes. 2017.
\newblock {Predicting the Outcome of Appeal Decisions in Germany's Tax Law}.
\newblock In \emph{Electronic Participation}, pages 89--99, Cham. Springer
  International Publishing.

\bibitem[{Warner and Hirschberg(2012)}]{Warner2012}
William Warner and Julia Hirschberg. 2012.
\newblock {Detecting Hate Speech on the World Wide Web}.
\newblock In \emph{Proceedings of the Second Workshop on Language in Social
  Media}, pages 19--26, Stroudsburg, PA, USA. Association for Computational
  Linguistics.

\bibitem[{Waseem et~al.(2017)Waseem, Davidson, Warmsley, and
  Weber}]{Waseem2017}
Zeerak Waseem, Thomas Davidson, Dana Warmsley, and Ingmar Weber. 2017.
\newblock {Understanding Abuse: A Typology of Abusive Language Detection
  Subtasks}.
\newblock In \emph{Proceedings of the First Workshop on Abusive Language
  Online}, pages 78--84. Association for Computational Linguistics.

\bibitem[{Wolf et~al.(2020)Wolf, Debut, Sanh, Chaumond, Delangue, Moi, Cistac,
  Rault, Louf, Funtowicz, Davison, Shleifer, von Platen, Ma, Jernite, Plu, Xu,
  Scao, Gugger, Drame, Lhoest, and Rush}]{Wolf2020}
Thomas Wolf, Lysandre Debut, Victor Sanh, Julien Chaumond, Clement Delangue,
  Anthony Moi, Pierric Cistac, Tim Rault, Rémi Louf, Morgan Funtowicz, Joe
  Davison, Sam Shleifer, Patrick von Platen, Clara Ma, Yacine Jernite, Julien
  Plu, Canwen Xu, Teven~Le Scao, Sylvain Gugger, Mariama Drame, Quentin Lhoest,
  and Alexander~M. Rush. 2020.
\newblock \href {https://www.aclweb.org/anthology/2020.emnlp-demos.6}
  {Transformers: State-of-the-art natural language processing}.
\newblock In \emph{Proceedings of the 2020 Conference on Empirical Methods in
  Natural Language Processing: System Demonstrations}, pages 38--45, Online.
  Association for Computational Linguistics.

\bibitem[{Xu et~al.(2012)Xu, Jun, Zhu, and Bellmore}]{Xu2012}
Jun-Ming Xu, Kwang-Sung Jun, Xiaojin Zhu, and Amy Bellmore. 2012.
\newblock \href {http://dl.acm.org/citation.cfm?id=2382029.2382139} {{Learning
  from Bullying Traces in Social Media}}.
\newblock In \emph{Proceedings of the Conference of the North American Chapter
  of the Association for Computational Linguistics: Human Language
  Technologies}, NAACL HLT '12, pages 656--666, Stroudsburg, PA, USA.
  Association for Computational Linguistics.

\bibitem[{Zufall et~al.(2019)Zufall, Horsmann, and Zesch}]{Zufall2019}
Frederike Zufall, Tobias Horsmann, and Torsten Zesch. 2019.
\newblock {From Legal to Technical Concept: Towards an Automated Classification
  of German Political Twitter Postings as Criminal Offenses }.
\newblock In \emph{Proceedings of the 2019 Conference of the North American
  Chapter of the Association for Computational Linguistics: Human Language
  Technologies, Volume 1 (Long Papers)}, NAACL HLT '19, pages 1337--1347.
  Association for Computational Linguistics.

\end{thebibliography}
\bibliographystyle{acl_natbib}

\end{document}